\documentclass[12pt, a4paper, conference, compsocconf, onecolumn]{IEEEtran}
\ifCLASSINFOpdf
\else
\fi
\usepackage{amsmath,epsfig}
\usepackage{amsmath}
\usepackage{epstopdf}
\usepackage{graphicx}
\usepackage{amsfonts}
\usepackage{multirow}
\usepackage{wrapfig}
\usepackage{subfloat}
\usepackage{lipsum}
\usepackage[lined,boxed,commentsnumbered]{algorithm2e}
\usepackage{color}
\usepackage{hyperref}
\usepackage{MnSymbol}

\usepackage{subfig}
\usepackage{makeidx}
\usepackage{moreverb}
\usepackage[lined,boxed,commentsnumbered]{algorithm2e}
\usepackage{array}
\usepackage{bbding}
\usepackage{booktabs}
  \DeclareGraphicsExtensions{.eps, .jpg,.png,.tif}

\begin{document}

\title{Semi-supervised Segmentation Fusion of Multi-spectral and Aerial Images} %

\author{\IEEEauthorblockN{Mete Ozay\IEEEauthorrefmark{1}}
\IEEEauthorblockA{\IEEEauthorrefmark{1}School of Computer Science \\
University of Birmingham \\
Edgbaston, Birmingham, B15 2TT, United Kingdom \\ Email: m.ozay@cs.bham.ac.uk}}

\maketitle

\begin{abstract}
A Semi-supervised Segmentation Fusion algorithm is proposed using consensus and distributed learning. The aim of Unsupervised Segmentation Fusion (USF) is to achieve a consensus among different segmentation outputs obtained from different segmentation algorithms by computing an approximate solution to the NP problem with less computational complexity. Semi-supervision is incorporated in USF using a new algorithm called Semi-supervised Segmentation Fusion (SSSF). In SSSF, side information about the co-occurrence of pixels in the same or different segments is formulated as the constraints of a convex optimization problem. The results of the experiments employed on artificial and real-world benchmark multi-spectral and aerial images show that the proposed algorithms perform better than the individual state-of-the art segmentation algorithms. 

\end{abstract}



\IEEEpeerreviewmaketitle

\section{Introduction} 
\label{chap:sf}

Image segmentation is one of the most important, yet unsolved problems in computer vision and image processing. Various segmentation algorithms studied in the literature have been applied to segment the objects in images \cite{for,gonz}. However, there are two main challenges of their employment. 

The first challenge is to \textit{extract} a robust structure, e.g. shape, of the segments by analyzing the outputs of segmentation algorithms when a target segmentation is not available with a training dataset. This challenge has been studied as a  \textit{segmentation mining} problem and analyzed as a consensus segmentation problem \cite{firfir,fl3} using an Unsupervised Segmentation Fusion approach by Ozay et al. \cite{sf}. The second challenge is the selection of an \textit{appropriate} algorithm with its parameters that provides an \textit{optimal} segmentation which is closer to a \textit{target} segmentation if a target segmentation is available with a training dataset. For this purpose, some of the segments in the segmentation set are expected to represent acquired target objects in the Unsupervised Segmentation Fusion algorithms \cite{firfir,sf,fl3}. In order to relax this assumption, first the error and distance functions of the algorithm should be refined to include these requirements. Therefore, prior information on the statistical properties of the datasets need to be incorporated using supervision. Then, side information about a target segmentation output should be used in the unsupervised segmentation fusion algorithm, which leads to a semi-supervised algorithm. In this work, this challenge has been analyzed by Semi-supervised Segmentation Fusion which incorporates prior and side information obtained from training datasets and expert knowledge to the USF algorithm \cite{sf}.


Consensus segmentation problem is re-formalized as a semi-supervised segmentation fusion problem and studied using decision fusion approaches \cite{o3} with semi-supervised learning \cite{semibook}. For this purpose, an algorithm called Semi-supervised Segmentation Fusion (SSSF) is introduced for fusing the segmentation outputs (decisions) of base-layer segmentation algorithms by incorporating the prior information about the data statistics and side-information about the content into the USF algorithm \cite{sf}. In the SSSF, this is accomplished by extracting the available side information about the targets, such as defining the memberships of pixels for the segments which represent a specific target in images. For this purpose, the side information about the pixel-wise relationships is reformulated and incorporated with a set of constraints in the segmentation fusion problem. In addition, a new distance function is defined for the Semi-supervised Segmentation Fusion by assigning weights to each segmentation. 

In order to compute the \textit{optimal} weights, the median partition (segmentation) problem is converted into a convex optimization problem. The side information which represents the pixel-wise segmentation membership relations defined by must-link and cannot-link constraints are incorporated in an optimization problem and in the structure of distance functions. Moreover, sparsity of the weights are used in the optimization problem for segmentation (decision) \textit{selection}. Various weighted cluster aggregation methods have been used in the literature \cite{w1,convex2}. Unlike these methods, the proposed approach and the algorithms enable learning both the structure of the distance function, the pixel-wise relationships and the \textit{contributions} of the decisions of the segmentation algorithms from the data by solving a single optimization problem using semi-supervision.

In the next section, a brief overview of USF algorithm is given. Semi-supervised Segmentation Fusion algorithm is introduced in Section \ref{sec:sssf}. Experimental analyses of the algorithms are given in Section \ref{sec:sssf_experiments}. Section \ref{sec:sf_conc} concludes the paper.

\section{Unsupervised Segmentation Fusion}

In the unsupervised segmentation fusion problem \cite{sf}, an image $\mathbb{I}$ is fed to $J$ different base-layer segmentation algorithms $SA_j$, $j=1,2, \ldots,J$. Each segmentation algorithm is employed on $\mathbb{I}$ to obtain a set of segmentation outputs $S_j= \{ s_{i} \} ^{n_j} _{i=1} $ where $s_i \in A ^N$ is a segmentation (partition) output, $A$ is the set of segment labels (names) with $N$ pixels, $|A|=C$ different segment labels, and a distance function $d(\cdotp, \cdotp)$. Note that $A ^N$ is the class of all segmentations of finite sets with $C$ different segment labels in the image $\mathbb{I}$.

An initial segmentation $s$ is selected from the segmentation set $S=\bigcup\limits ^J _{j=1} S_j$ consisting of $K=\sum \limits ^J _{j=1} n_j$ segmentations using algorithms which employ search heuristics, such as Best of K \textit{(BOK)} \cite{BOEM}. Then, a \textit{consensus} segmentation $\hat { s }$ is computed by solving the following optimization problem:
\[
\hat { s } = \underset{ s }{\mathrm{argmin}} \sum\limits ^{K} _{i=1} d( s _i , s ) \; . 
\]
Given two segmentations $s_i$ and $s_j$, the distance function is defined as the \textit{Symmetric Distance Function (SDD)} given by $d( s_i , s_j)=N_{01} + N_{10}$, where $N_{01}$ is the number of pairs co-segmented in $s_i$ but not in $s_j$, and $N_{10}$ is the number of pairs co-segmented in $s_j$ but not in $s_i$ \cite{BOEM}. 

This optimization problem was solved by Ozay et al. \cite{sf} using an Unsupervised Segmentation Fusion algorithm. At each iteration $t$ of the optimization algorithm, a new segmentation is computed. Specifically, using the assumption that single element updates do not change the objective function $H_t = \sum\limits ^{K} _{i=1} d( s _i , s_t )$, $H_t$ is approximated by $H_{t-1}$ with a scale parameter $ \beta \in [0,1] $. Then, the current \textit{best one element move} is updated at $t$ using
\[
\Delta s _t = \frac{ \partial } { \partial s _t } ( \beta H_{t-1} + d( s_{i'}, s_t) ) \; ,
\]
where $s_{i'}$ is the randomly selected segmentation. If an $N \times C$ matrix $[H]$ is defined such that the $n ^{th}$ row and the $c ^{th}$ column of the matrix, $[H] _{nc}$, is the updated value of $H$ obtained by switching $n ^{th}$ element of $s$ to the $c ^{th}$ segment label, then the move can be approximated by
\begin{equation}
\underset{ n,c}{\mathrm{argmin}} \; \beta [H_{t-1} ]_{n,c} + [d( s _{i'} , s _t )]_{n,c} \; , 
\label{eq:app}
\end{equation}
if $s_{i'}$ is selected for updating $s _t$ at time $t$, $\forall i=1,2,\ldots,N$, $\forall c=1,2,\ldots,C$. If there is no improvement on the best move or a termination time $T$ is achieved, the current segmentation is returned by the USF algorithm \cite{sf}.

\section{Incorporating Prior and Side Information to Segmentation Fusion} 
\label{sec:sssf}

In this section, we introduce a new Semi-supervised Segmentation Fusion algorithm which solves weighted decision and distance learning problems that are mentioned in Section \ref{chap:sf} by incorporating side-information about the pixel memberships into the unsupervised Segmentation Fusion algorithm. Then, the goal of the proposed Semi-supervised Segmentation Fusion algorithm can be summarized as obtaining a segmentation which is close to both base-layer segmentations and a target segmentation using weighted distance learning and semi-supervision.

In the weighted distance learning problem, some of the weights may be required to be zero, in other words, sparsity may be required in the space of weight vectors to select the decision of some of the segmentation algorithms. For instance, if fusion is employed on multi-spectral images with large number of bands, and if some of the most informative bands are needed to be selected, then sparsity defined by the weight vectors becomes a very important property. In addition, side information about the pixel-wise relationships of the segmentations can be defined in distance functions. Thereby, both the structure of the distance function, the pixel-wise relationships and the \textit{contributions} of the decisions of the segmentation algorithms can be learned from the data. 

\subsection{Formalizing Semi-supervision for Segmentation Fusion}
\label{sec:sssf_problem}

We define Semi-supervised Segmentation Fusion problem as a convex constrained stochastic sparse optimization problem. In the construction of the problem, first pixel-wise segment memberships are encoded in the definition of a semi-supervised weighted distance learning problem by decomposing Symmetric Distance Function ($SDD$) as \cite{w1}
\begin{equation}
d(s_i, s_j)=\sum ^N _{m=1} \sum ^N _{l=1} d_{m,l}(s_i, s_j),
\label{eq:sssf_df}
\end{equation}
and
\[
d_{m,l}(s_i, s_j) =
\begin{cases}
1, & \mathrm{if} \; (m,l) \in \Theta_c(s_i) \; \mathrm{and} \; (m,l) \notin \Theta_c(s_j) \\ 
1, & \mathrm{if} \; (m,l) \notin \Theta_c(s_i) \; \mathrm{and} \; (m,l) \in \Theta_c(s_j) \\ 
0, & \mathrm{otherwise}
\end{cases} ,
\]
where $(m,l) \in \Theta_c(s_i)$ means that the pixels $m$ and $l$ belong to the same segment $\Theta_c$ in $s_i$ and $(m,l) \notin \Theta_c(s_i)$ means that $m$ and $l$ belong to different segments in $s_i$. Then, a connectivity matrix $M$ is defined with the following elements;

\begin{equation}
M_{ml} (s_i) = 
\left\{ \begin{array}{rcl}
1, & \mbox{if} \; (m,l) \in \Theta_c(s_i) \\ 
0, & \mathrm{otherwise}
\end{array}\right .
\label{eq:m_def}
\end{equation}

Note that \cite{w1}, 
\begin{equation}
d_{m,l}(s_i, s_j)= [ M_{m,l}(s_i)-M_{m,l}(s_j) ]^2 \;.
\label{eq:m}
\end{equation}


Then, the distance between the connectivity matrices of two segmentations $s$ and $s_i$ is defined as \cite{convex2}
\begin{equation}
d_{\kappa}(M(s),M(s_i)) = \sum ^N _{m=1} \sum ^N _{l=1} d_{\kappa}(M_{m,l}(s),M_{m,l}(s_i)) \; ,
\label{eq:dm}
\end{equation}
where $d_{\kappa}$ is the Bregman divergence defined as
\[
d_{\kappa}(x,y)=\kappa(x)-\kappa(y)-\nabla \kappa(y)(x-y) \; ,
\]
and $\kappa : \mathbb{R} \rightarrow \mathbb{R} $ is a strictly convex function. Since $d_{\kappa}$ is defined in \eqref{eq:m} as Euclidean distance, \eqref{eq:dm} is computed during the construction of best one element moves. 

In order to compute the weights of base-layer segmentations during the computation of distance functions, the following quadratic optimization problem is defined;
\begin{eqnarray}
\underset{ \bar{w} }{\mathrm{argmin}} \; \sum\limits ^{ K} _{i=1} w_i d_{\kappa}(M(s),M(s_i)) + \lambda_q \parallel \bar{w} \parallel ^2 _2 \nonumber \\
\mathrm{s.t.} \sum\limits ^{ K} _{i=1} w_i = 1, w_i \geq 0, \forall i=1,2,\ldots,K \; ,
\label{eq:qp}
\end{eqnarray}
where $\lambda_q >0$ is the regularization parameter and $\bar{w} = (w_1, w_2,\ldots,w_K)$ is the weight vector. Since we use $\sum\limits ^{ K} _{i=1} w_i = 1$ and $w_i \geq 0$ in the constraints of the optimization problem \eqref{eq:qp}, we enable the \textit{selection} and \textit{removal} of a base-layer segmentation $s_i$ by assigning $w_i=0$ to $s_i$.

Defining the distance function \eqref{eq:sssf_df} in terms of the segment memberships of the pixels \eqref{eq:m_def} in \eqref{eq:m}, must-link and cannot-link constraints can be incorporated to the constraints of \eqref{eq:qp} as follows;
\begin{equation}
M_{ml} (s_i)= 
\begin{cases}
1, & \mathrm{if} (m,l) \in \mathfrak{M} \\
0, & \mathrm{if} (m,l) \in \mathfrak{C}
\end{cases} 
,
\label{eq:mc}
\end{equation}
where $\mathfrak{M}$ is the set of must-link constraints and $\mathfrak{C}$ is the set of cannot-link constraints. Then, the following optimization problem is defined for Semi-supervised Segmentation Fusion
\begin{eqnarray}
\underset{ M(s) }{\mathrm{argmin}} \; \sum\limits ^{ K} _{i=1} d_{\kappa}(M(s),M(s_i)) + \lambda_q \parallel \bar{w} \parallel ^2 _2 \nonumber \\
\mathrm{s.t} \; \; \; \; M_{ml} (s_i)=1, \mathrm{if} (m,l) \in \mathfrak{M} \nonumber \\
M_{ml} (s_i)=0, \mathrm{if} (m,l) \in \mathfrak{C} \; .  
\label{eq:ssc}
\end{eqnarray}

Wang, Wang and Li \cite{convex2} analyze generalized cluster aggregation problem using \eqref{eq:ssc} for fixed weights $\bar{w}$ and define the solution set as follows;
\begin{enumerate}
\item If $(m,l) \in \mathfrak{M}$ or $(m,l) \in \mathfrak{C}$, then \eqref{eq:mc} is the solution set for $(k,l)$,
\item If $(m,l) \notin \mathfrak{M}$ and $(m,l) \notin \mathfrak{C}$, then $M_{ml} (s_i)$ can be solved by
\[
\nabla_{\kappa} M_{ml} (s_i)= \sum\limits ^{ K} _{i=1} w_i \nabla_{\kappa} (M(s_i)).
\]
\end{enumerate}
Then, they solve \eqref{eq:qp} for fixed $M (s)$. Note that, $\ell_2$ norm regularization does not assure sparsity efficiently \cite{lasso} because $\parallel \bar{w} \parallel ^2 _2$ is a quadratic function of the weight variables $w_i$ which treats each $w_i$ equally. In order to control the sparsity of the weights by treating each $w_i$ different from the other weight variables $w_{j \neq i}$ using a linear function of $w_i$, such as $\parallel \bar{w} \parallel _1$ which is the $\ell_1$ norm of $\bar{w}$, a new optimization problem is defined as follows;
\begin{eqnarray}
\underset{( M(s), \bar{w} ) }{\mathrm{argmin}} \; \sum\limits ^{ K} _{i=1} w_i d_{\kappa}(M(s),M(s_i)) + \lambda \parallel \bar{w} \parallel _1 \nonumber \\
\mathrm{s.t} \; \; \; \sum\limits ^{ K} _{i=1} w_i =1, w_i \geq 0, \forall i=1,2,\ldots,K \nonumber \\
M_{ml} (s_i)=1, \mathrm{if} (m,l) \in \mathfrak{M} \nonumber \\
M_{ml} (s_i)=0, \mathrm{if} (m,l) \in \mathfrak{C} \; ,
\label{eq:lasso}
\end{eqnarray}
where $\lambda \in \mathbb{R}$ is the parameter which defines the sparsity of $\bar{w}$. Similarly, \eqref{eq:lasso} is computed in two parts;
\begin{enumerate}

\item For fixed $M(s)$, solve
\begin{eqnarray}
\underset{ \bar{w} }{\mathrm{argmin}} \; \sum\limits ^{ K} _{i=1} w_i d_{\kappa}(M(s),M(s_i)) + \lambda \parallel \bar{w} \parallel _1 \nonumber \\
\mathrm{s.t} \; \; \; \sum\limits ^{ K} _{i=1} w_i =1, w_i \geq 0, \forall i=1,2,\ldots,K \; .
\label{eq:lasso1}
\end{eqnarray}

\item For fixed $\bar{w}$, solve
\begin{eqnarray}
 \underset{ M(s) }{\mathrm{argmin}} \; \sum\limits ^{ K} _{i=1} w_i d_{\kappa}(M(s),M(s_i)) + \lambda \parallel \bar{w} \parallel _1 \nonumber \\
 \mathrm{s.t} \; \; \; \; M_{ml} (s_i)=1, \mathrm{if} (m,l) \in \mathfrak{M} \nonumber \\
 M_{ml} (s_i)=0, \mathrm{if} (m,l) \in \mathfrak{C} \; .
\label{eq:lasso2}
\end{eqnarray}
\end{enumerate}

An algorithmic description of Semi-supervised Segmentation Fusion which solves \eqref{eq:lasso1} and \eqref{eq:lasso2} is given in the next subsection.

\subsection{Semi-supervised Segmentation Fusion Algorithm}

In the proposed Semi-supervised Segmentation Fusion algorithm, \eqref{eq:lasso1} and \eqref{eq:lasso2} are solved to compute weighted distance functions which are used in the construction of best one element moves. 

In Algorithm \ref{alg:sssf}, first the weight vector $\bar{w}$ is computed by solving \eqref{eq:lasso1} for each selected segmentation $s_{i'}$ in the $4^{th}$ step of the algorithm. In order to solve \eqref{eq:lasso1} using an optimization method called Alternating Direction Method of Multipliers (ADMM) \cite{admm}. ADMM has been employed to solve \eqref{eq:lasso1} until a termination criterion $\tau \leq T_{\tau}$ or convergence is achieved \cite{admm}. Once the weight vector $\bar{w}$ is computed in the $4^{th}$ step, \eqref{eq:lasso2} is solved in the $5^{th}$, $6^{th}$ and $7^{th}$ steps of the algorithm: $\bar{w} d( s_{i'}, s) + \lambda \parallel \bar{w} \parallel _1$ is computed using $M(s_{i'})$ and $\bar{w}$ in the $5^{th}$ step, $ [H_t]$ is computed in the $6^{th}$ step and  $\Delta s$ is computed in the $7^{th}$ step to update $s$. Note that the sparse weighted distance function, which is approximated by $\beta [H_t] + [\bar{w} d( s_{i'}, s) +\lambda \parallel \bar{w} \parallel _1]$ in Algorithm \ref{alg:sssf}, is different from the distance function in USF. In addition, each segmentation is selected sequentially in a pseudo-randomized permutation order in Algorithm \ref{alg:sssf}. If an initially selected segmentation performs better than the other segmentations, then the algorithm may be terminated in the first running over the permutation set. Otherwise, the algorithm runs until the termination time $T$ is achieved or all of the segmentations are selected.  

\begin{algorithm}
%
\SetKwFunction{Union}{Union}\SetKwFunction{FindCompress}{FindCompress}
\SetKwInOut{Input}{input}\SetKwInOut{Output}{output}
\Input {Input image $I$, $\{ SA_j \} ^J _{j=1}$, $T$, $T_{\tau}$. }
\Output{Output segmentation $O$.}
\nl Run $SA_j$ on $I$ to obtain $S_j= \{ s_{i} \} ^{u_j} _{i=1}$, $\forall j=1,2,\ldots,J.$\;
\nl At $t=1$, initialize $s$ and $ [H_t]$\;
\For{$t \leftarrow 2$ \KwTo $T$}{
\nl Randomly select one of the segmentation results with an index $i' \in \{ 1,2,\ldots,K \}$\;
\nl Solve \eqref{eq:lasso1} for $M(s_{i'})$ to compute $w_k$\;
\nl Compute $\bar{w} d( s_{i'}, s) + \lambda \parallel \bar{w} \parallel _1$\;
\nl $ [H_t] \leftarrow \beta [H_t] + [\bar{w} d( s_{i'}, s) +\lambda \parallel \bar{w} \parallel _1]$\;
\nl Compute $\Delta s$ by solving $\underset{ n,c }{\mathrm{argmin}} \beta [H_{t} ]_{n,c}$ \;
\nl $ s \leftarrow s + \Delta s$\; 
\nl $ t \leftarrow t+1$\; 
}
\nl$ O \leftarrow s$ \;
\caption{Semi-supervised Segmentation Fusion.}
\label{alg:sssf}
\end{algorithm}

\section{Experiments}                                                    
\label{sec:sssf_experiments}

In this section, the proposed Semi-supervised Segmentation Fusion (SSSF) algorithm is analyzed on real world benchmark multi-spectral and aerial images \cite{multispec,porway,lhi}. In the implementations, three well-known segmentation algorithms, $k$-means, Mean Shift cite{ms} and Graph Cuts \cite{gc1,cut_seg,gc3} are used as the base-layer segmentation algorithms. Three indices are used to measure the performances between the output images $O$ and the ground truth of the images: i) Rand Index ($RI$), ii) Adjusted Rand Index ($ARI$), and iii) Adjusted Mutual Information ($AMI$) \cite{Info_theoretic} which adjusts the effect of mutual information between segmentations due to chance, similar to the way the $ARI$ corrects the $RI$.

In the experiments, a Graph Cut implementation of Veksler \cite{gc3} for image segmentation is used with a Matlab wrapper of Bagon \cite{gc1} and the source code provided by Shi \cite{nc1}. The algorithm parameters are selected by first computing $ARI$ values between a given target segmentation and each segmentation computed for each parameter $ \sigma_r \in \{ 0.1, 0.2, \ldots, 10 \}$, $ \sigma_s \in \{ 1, 2, \ldots, 100 \}$, $ r_{ncut} \in \{ 1, 2, \ldots, 100 \}$ and $\tau_{ncut} \in \{ 0.01, 0.02, \ldots, 1 \}$ \cite{nc1}. Then, a parameter 4-tuple $(\sigma_r, \sigma_s, r_{ncut}, \tau_{ncut})$ which maximizes $ARI$ is selected\footnote{Minimization of $ARI$ is considered as the cost function in the estimation of parameters following the relationship between $ARI$ and $SDD$ as well as it is one of the performance measures \cite{sf}.}. Similarly, a parameter 3-tuple $(h_s, h_r, mA)$ which maximizes $ARI$ is selected for Mean Shift algorithm from the parameter sets $ h_s \in \{ 1,3, 5,10,50,100 \}$, $ h_r \in \{ 1,3, 5,10,50,100 \}$ and $ mA \in \{ 100,200 \ldots, 10000 \}$ \cite{ms}. For $k$-means, $k=C$ is used, if not stated otherwise. Assuming that $C$ is not known in the image, a parameter search algorithm proposed in \cite{sf} is employed using the training data in order to find the optimal $\hat{C}$ for $c=2,3,4,5,6,7,8,9,10$. Similarly, the parameter estimation algorithm suggested in \cite{sf} is employed for a set of $\hat{\beta}$ values $\Xi= \{ 0.1, 0.2, 0.3, 0.4, 0.5, 0.6, 0.7, 0.8, 0.9, 0.99 \}$. 

The termination parameter of SSSF and ADMM is taken as $T=1000$ and $T_{\tau}=1000$, respectively. The penalty parameter of ADMM is chosen as $\theta=1$ as suggested in \cite{admm}. The regularization parameter is computed as $\lambda=0.5 \lambda_{max}$ \cite{admm}, where $ \lambda_{max}=max \{ ||d_{\kappa}(M(s),M(s_a)) \bar{y} || _2 \}_{a=1}^K$, $y_n=||d_{\kappa}(M(s),M(s_{n})) \bar{w} || _2$, $S=\{ s_{n} \}_{n=1} ^N $ is the set of segments in an training image and $\bar{y}=[y_1,y_2,\ldots,y_N]$ is the labels of segments in $S$. Then, $\lambda$ is computed in training phase and employed in both training and test phases. In the training phase, $\lambda$ and $\bar{w}$ are computed, and  the constraints $\mathfrak{M}$ and $\mathfrak{C}$ are constructed using the ground truth data, i.e. pixel labels of training images as described in Section \ref{sec:sssf_problem}. In the testing phase, \eqref{eq:m_def} is employed for the construction of connectivity matrices and  $[\bar{w} d( s_ik, s) +\lambda \parallel \bar{w} \parallel _1]$ is computed $\forall i=1,2,\ldots,K$. The performance of the proposed SSSF is compared with the performances of $k$-means, Mean Shift, Graph Cuts, Unsupervised Segmentation Fusion (USF) \cite{sf}, Distance Learning (DL) \cite{sf} and Quasi-distance Learning (QD) \cite{sf} algorithms.

\subsection{Analyses on Multi-spectral Images}

\begin{figure}[htbp]
\centering

\subfloat[Training dataset.]
{\includegraphics[scale=0.5]{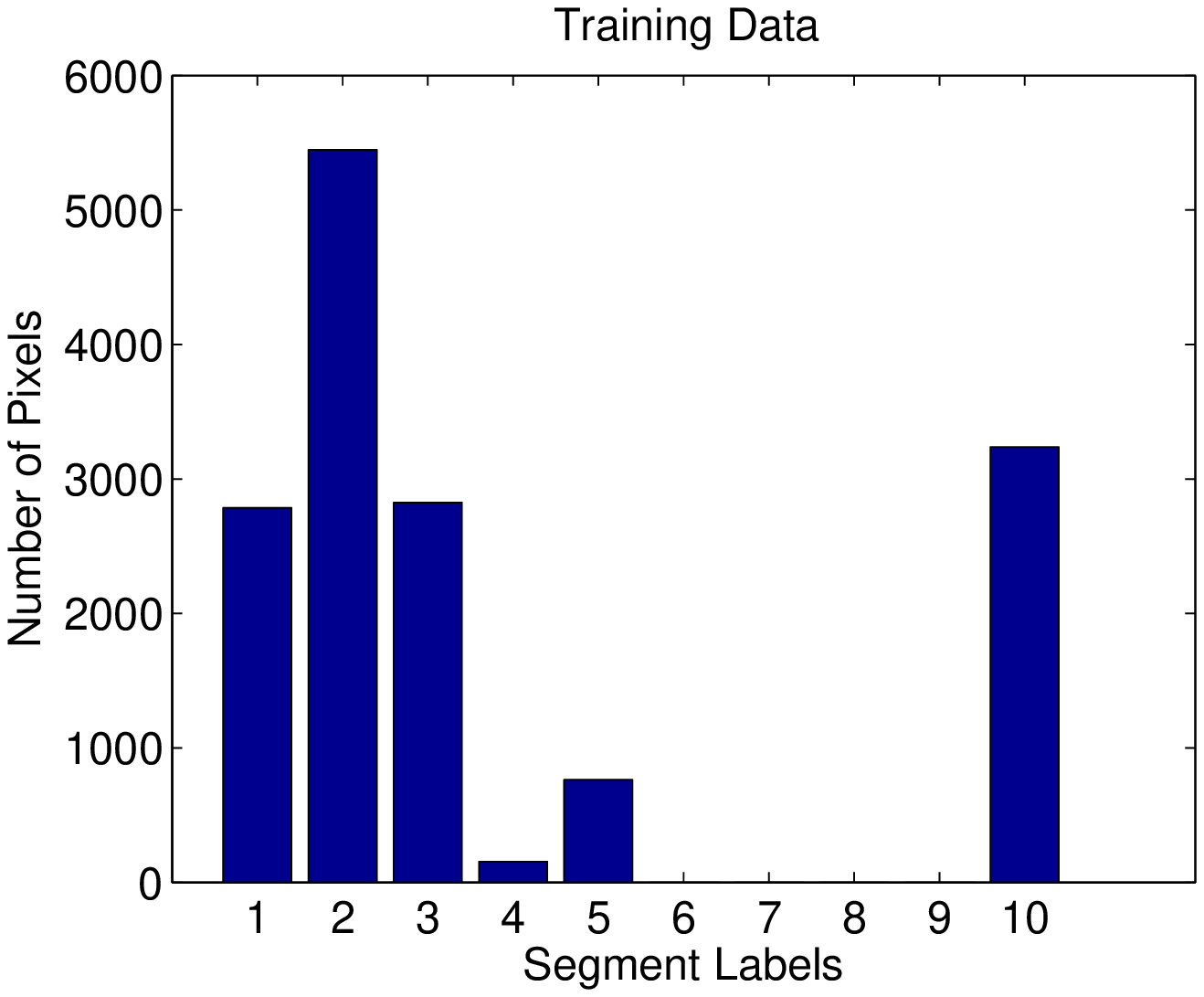}} 
\subfloat[Test dataset.]{\includegraphics[scale=0.5]{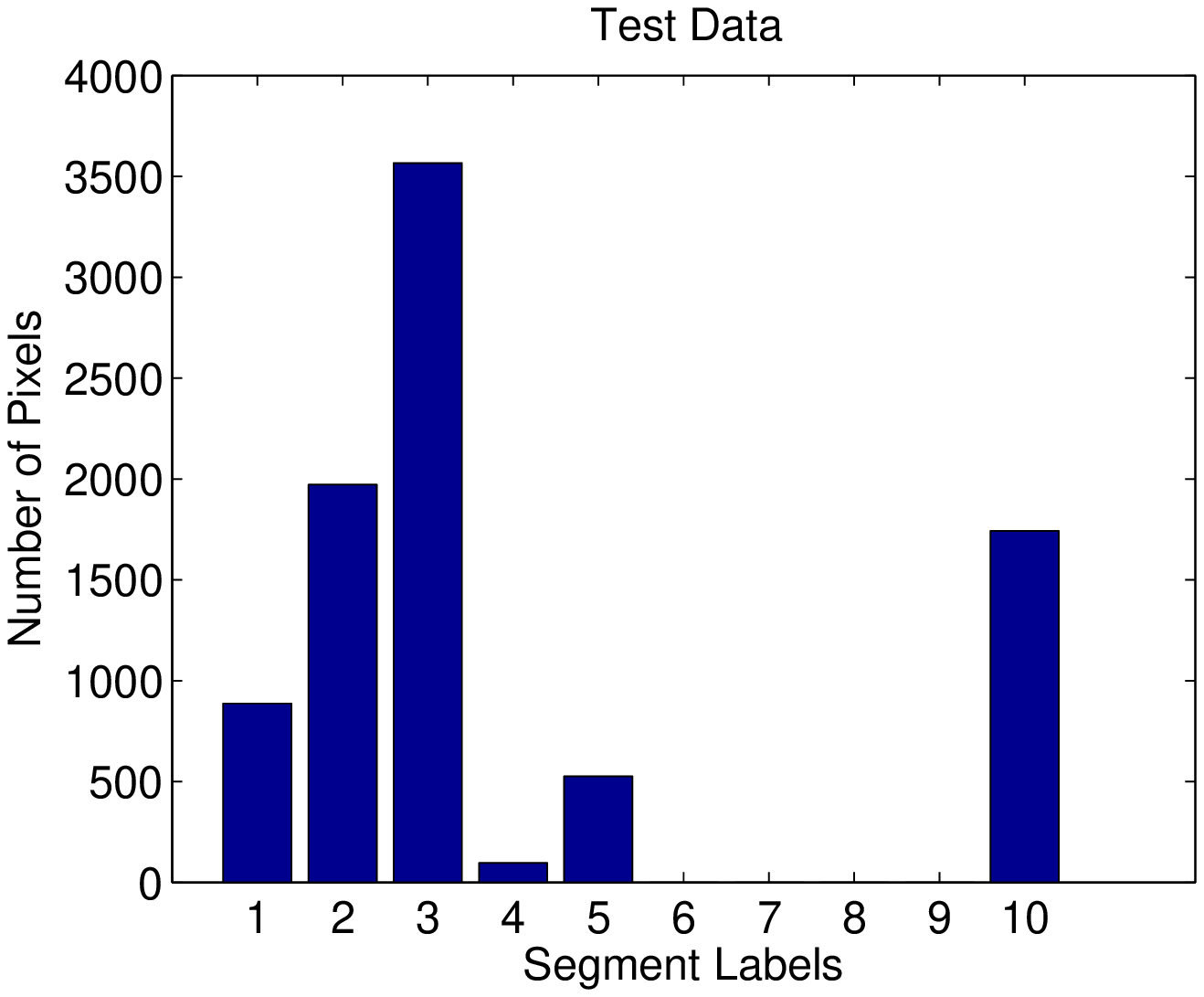}} 

\caption{Distribution of pixels given the segment labels in TMI.}
\label{fig:9}
\end{figure}

In the first set of experiments, the proposed algorithms are employed on $7$ band Thematic Mapper Image (TMI) which is provided by MultiSpec \cite{multispec}. The image with size $169 \times 169$ is split into training and test images: i) a subset of the pixels with coordinates $x=(1:169)$ and $y=(1:90)$ is taken as the training image and ii) a subset of the pixels with coordinates $x=(1:169)$ and $y=(91:142)$ is taken as the test image. Dataset is split in order to obtain segments with at least $100$ pixels both in training and test images. In the images, there are $C=6$ number of different segment labels. The distribution of pixels given the segment labels is shown in Fig.~\ref{fig:9}.

\begin{table*}[htbp]
\centering
\caption{Training and test performances of the algorithms for Thematic Mapper Image.}
\begin{tabular}{ccccccccccc}
\addlinespace
\toprule
& \multicolumn{2}{c}{\textbf{Average Base}} & \multicolumn{2}{c}{\textbf{USF}} & \multicolumn{2}{c}{\textbf{DL}} & \multicolumn{2}{c}{\textbf{QD}} & \multicolumn{2}{c}{\textbf{SSSF}} \\
\midrule
& \textbf{Tr} & \textbf{Te} & \textbf{Tr} & \textbf{Te} & \textbf{Tr} & \textbf{Te} & \textbf{Tr} & \textbf{Te} & \textbf{Tr} & \textbf{Te}\\
\textbf{RI} & 0.730 & 0.703 & 0.731 & 0.704 & 0.738 & 0.710& 0.732 & 0.714 & 0.792 & 0.740 \\
\textbf{ARI} & 0.264 & 0.159 & 0.265 & 0.160 & 0.282 & 0.184 & 0.270 & 0.174 & 0.305 & 0.220 \\
\textbf{AMI} & 0.182 & 0.187 & 0.182 & 0.188 & 0.205 & 0.203 & 0.198 & 0.204 & 0.251 & 0.237\\
\bottomrule
\end{tabular}%
\label{tab:sssf_tab11}%
\end{table*}%

First $k$-means is implemented on different bands $\mathbb{I}_j$ of the multi-spectral image $\mathbb{I}=(\mathbb{I}_1, \mathbb{I}_2, \ldots, \mathbb{I}_J)$ for $J=7$, in order to perform multi-modal data fusion of different spectral bands using segmentation fusion. The results of the experiments on Thematic Mapper Image is given in Table \ref{tab:sssf_tab11}. In the Average Base column, the performance values of $k$-means algorithm averaged over $7$ bands are given. It is observed that the performance values of USF are similar to the arithmetic average of the performance values of $k$-means algorithms. When semi-supervision is used, a remarkable increase is observed in the performances in SSSF. However, full performance ($1$ values for the indices) is not achieved in training. Since the output image $O$ may not converge to the GT of the image, the convergence assumption mentioned in the previous section may not be valid for this image.

\begin{table*}[htbp]
\centering
\caption{Experiments on 7-band images.}
\begin{tabular}{ccccccccccccccc}
\addlinespace
\toprule
& \multicolumn{2}{c}{\textbf{$k$-means}} & \multicolumn{2}{c}{\textbf{Graph Cut}} & \multicolumn{2}{c}{\textbf{Mean Shift}} & \multicolumn{2}{c}{\textbf{USF}} & \multicolumn{2}{c}{\textbf{DL}} & \multicolumn{2}{c}{\textbf{QD }} & \multicolumn{2}{c}{\textbf{SSSF}} \\
\midrule
& \textbf{Tr} & \textbf{Te} & \textbf{Tr} & \textbf{Te} & \textbf{Tr} & \textbf{Te} & \textbf{Tr} & \textbf{Te} & \textbf{Tr} & \textbf{Te} & \textbf{Tr} & \textbf{Te} &  \textbf{Tr} & \textbf{Te} \\
\textbf{RI} & 0.742 & 0.715 & 0.754 & 0.717 & 0.710 & 0.714 & 0.711 & 0.714 & 0.713 & 0.710 & 0.752 & 0.724 & 0.801 & 0.733 \\
\textbf{ARI} & 0.167 & 0.125 & 0.234 & 0.132 & 0.266 & 0.176 & 0.267 & 0.176  & 0.270 & 0.180 & 0.262 & 0.178 & 0.326 & 0.236 \\
\textbf{AMI} & 0.176 & 0.183 & 0.193 & 0.190 & 0.195 & 0.209 & 0.196 & 0.209 & 0.195 & 0.205 & 0.198 & 0.211 & 0.220 & 0.219 \\
\bottomrule
\end{tabular}%
\label{tab:sssf_tab21}%
\end{table*}%

In the second set of the experiments, $k$-means, Graph Cut and Mean Shift algorithms are employed on $7$-band training and test images. Now, the image segmentation problem is considered as a pixel clustering problem in $7$ dimensional spaces. The results are given in Table \ref{tab:sssf_tab21}. The performance values of USF are closer to the performance values of the Mean Shift algorithm, since the output image of USF is closer to the output segmentation of the  Mean Shift algorithm. Moreover, SSSF provides better performance than the other algorithms, since  SSSF incorporate prior information by assigning higher weights to the partitions with higher performances. 

\begin{figure}[htbp]
\centering

\subfloat[Training dataset.]{\includegraphics[scale=0.5]{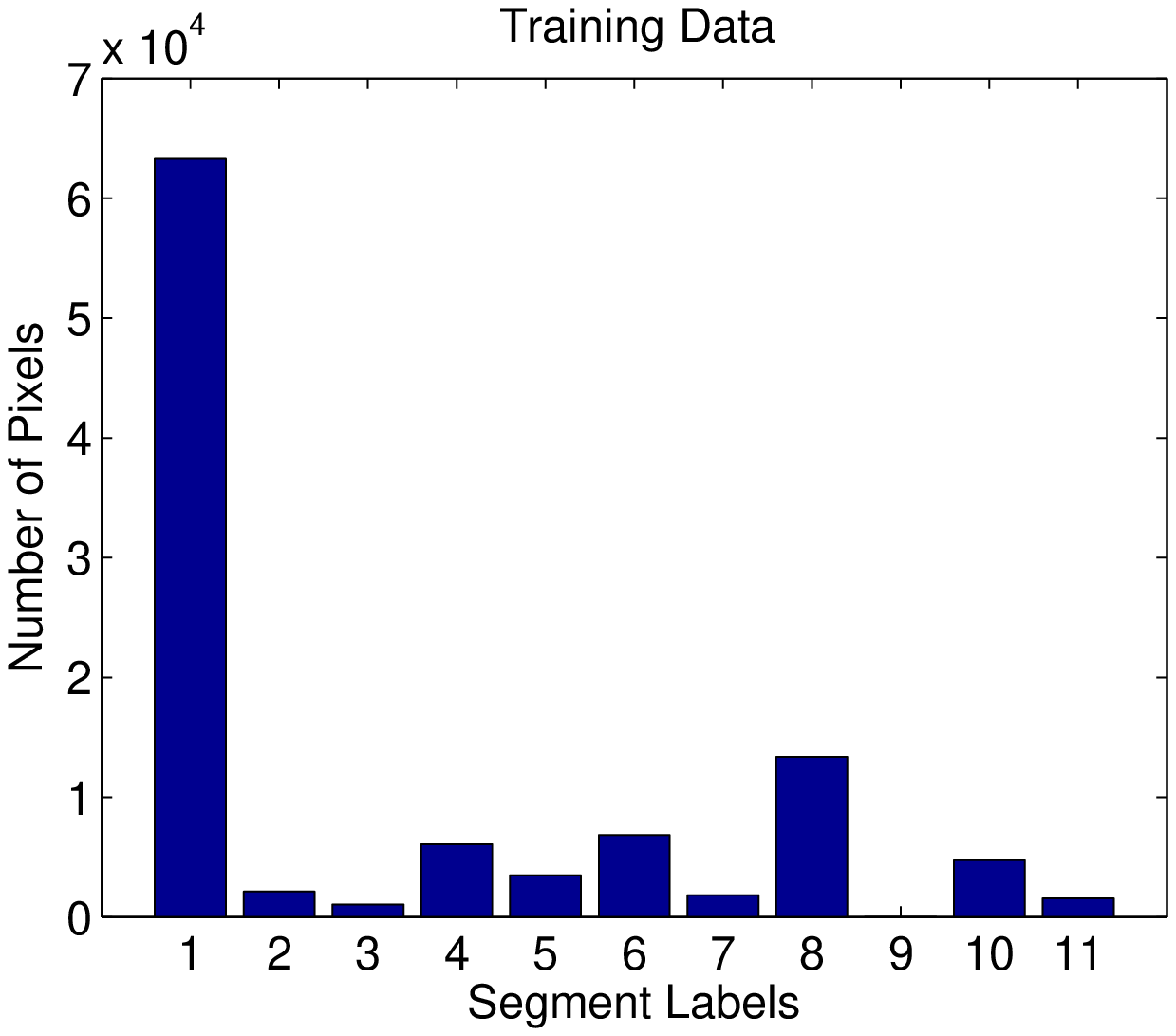}} 
\subfloat[Test dataset.]{\includegraphics[scale=0.5]{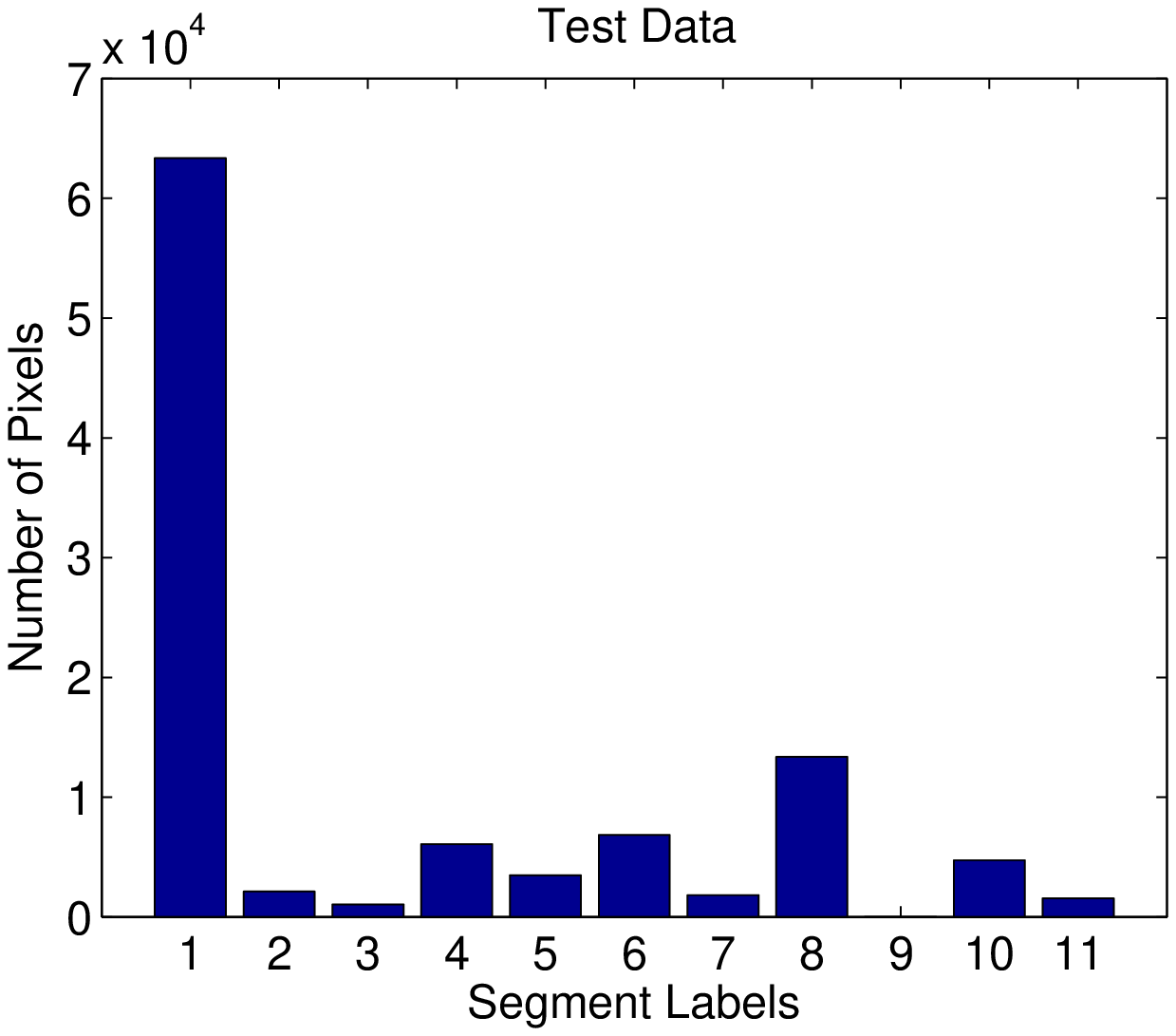}} 

\caption{Distribution of pixels given the segment labels in MDI.}
\label{fig:10}
\end{figure}

In the third set of experiments, $k$-means algorithm is employed on each band of $12$-band Moderate Dimension Image \cite{multispec}. The size of the image is $ 949 \times 220$, and there are $11$ segments in the GT of the image \cite{multispec}. The classes are background, Alfalfa, Br Soil, Corn, Oats, Red Cl, Rye, Soybeans, Water, Wheat, Wheat2. $104392$ pixels are randomly selected for training and the remaining $104388$ pixels are randomly selected for testing. In order to conserve the spatial distribution of the selected pixels, the pixels which reside in a segment with the same label in a spatial neighborhood are selected as test and training data. The distributions of pixels in training and test datasets are shown in Fig.~\ref{fig:10}. The results on the test data are given in Table \ref{tab:mod21}. It is observed that the performance values for USF are smaller than the average performance values of base-layer segmentation outputs. When prior information is employed using SSSF, it is observed that the smaller weights are assigned to the segmentations with relatively small performance values. In addition, the output images of SSSF are closer to the target segmentations obtained from the GT images. In summary, remarkable performance increases are observed in SSSF algorithm. 

\begin{table*}[htbp]
\centering
\caption{Performance of the algorithms for Moderate Dimension Image.}
\begin{tabular}{ccccccccccc}
\addlinespace
\toprule
& \multicolumn{2}{c}{\textbf{Average Base}} & \multicolumn{2}{c}{\textbf{USF}} & \multicolumn{2}{c}{\textbf{DL}} & \multicolumn{2}{c}{\textbf{QD}} & \multicolumn{2}{c}{\textbf{SSSF}}\\
\midrule
& \textbf{Tr} & \textbf{Te} & \textbf{Tr} & \textbf{Te} & \textbf{Tr} & \textbf{Te} & \textbf{Tr} & \textbf{Te} & \textbf{Tr} & \textbf{Te} \\
\textbf{RI} & 0.533 & 0.532 & 0.532 & 0.530 & 0.533 & 0.533& 0.535 & 0.530 & 0.553 & 0.550 \\
\textbf{ARI} & 0.008 & 0.009 & 0.007 & 0.007 & 0.013 & 0.011 & 0.010 & 0.011 & 0.109 & 0.110 \\
\textbf{AMI} & 0.139 & 0.141 & 0.124 & 0.120 & 0.123 & 0.121 & 0.123 & 0.124 & 0.177 & 0.185\\
\bottomrule
\end{tabular}%
\label{tab:mod21}%
\end{table*}%

\subsection{Analyses on Aerial Images}

In this section, the segmentation of roads in the aerial images is considered, which are analyzed in \cite{lhi}. Detailed information about the images in the dataset is given in \cite{porway,lhi}. $7$ training and $7$ test images with road and background labels are randomly selected from the dataset. The id numbers of the training and test images in the dataset are $tr=\{7, 26, 40, 41, 42, 43, 77\}$, and $te=\{78, 90, 91, 92, 93, 94, 95\}$, respectively. In order to observe the affect of the statistical similarity between training and test datasets, the performances are not averaged for different implementations of algorithms on random permutations of training and test images, and both of training and test performances are given in the results.

The results are shown in Table \ref{tab:road1}. It is observed that the performance indices of USF are the same as the indices of Mean Shift. This is basically because of the fact that Mean Shift has a higher number of different segment labels than the other algorithms. Therefore, the outputs of Mean Shift suppress the outputs of other algorithms in the computation of distance functions. Moreover, higher performances than the base-layer segmentation algorithms are obtained, when semi-supervision (SSSF) is employed in segmentation fusion.

\begin{table*}[htbp]
\centering
\caption{Performances of algorithms on Road Segmentation Dataset.}
\begin{tabular}{ccccccccccccccc}
\addlinespace
\toprule
& \multicolumn{2}{c}{\textbf{$k$-means}} & \multicolumn{2}{c}{\textbf{Graph Cut}} & \multicolumn{2}{c}{\textbf{Mean Shift}} & \multicolumn{2}{c}{\textbf{USF}} & \multicolumn{2}{c}{\textbf{DL}} & \multicolumn{2}{c}{\textbf{QD }}  & \multicolumn{2}{c}{\textbf{SSSF}} \\
\midrule
 & \textbf{Tr} & \textbf{Te} & \textbf{Tr} & \textbf{Te} & \textbf{Tr} & \textbf{Te} & \textbf{Tr} & \textbf{Te} & \textbf{Tr} & \textbf{Te} & \textbf{Tr} & \textbf{Te} & \textbf{Tr} & \textbf{Te}  \\
 \textbf{RI} & 0.513 & 0.535 & 0.512 & 0.523 & 0.379 & 0.328 & 0.378 & 0.328 & 0.392 & 0.353 & 0.407 & 0.390  & 0.550 & 0.563\\
\textbf{ARI} & 0.014 & 0.002 & 0.017 & 0.008 & 0.010 & 0.008 & 0.010 & 0.008 & 0.010 & 0.008 & 0.011 & 0.007 & 0.020 & 0.015\\
\textbf{AMI} & 0.404 & 0.003 & 0.054 & 0.006 & 0.053 & 0.070 & 0.044 & 0.070 & 0.082 & 0.080 & 0.090 & 0.080 & 0.422 & 0.110 \\
\bottomrule
\end{tabular}%
\label{tab:road1}%
\end{table*}%


\section{Conclusion}
\label{sec:sf_conc}
An algorithm called Semi-supervised Segmentation Fusion (SSSF) is introduced for fusing the segmentation outputs (decisions) of base-layer segmentation algorithms by incorporating the prior information about the data statistics and side-information about the content into the Unsupervised Segmentation Fusion algorithm. The proposed SSSF algorithm reformulates the segmentation fusion problem as a constrained optimization problem, where the constraints are defined in such a way to semi-supervise the segmentation process.

Experimental results show that the difference between $RI$ and $ARI$ values increases, as the number of segmentation outputs $K$ increases for a fixed number of segments $C$. We observe that one of the reasons for the observation of this fluctuation is the early termination of the USF and the proposed SSSF before a consensus segmentation is obtained. In addition, the performances of the base-layer segmentation algorithms and the proposed segmentation fusion algorithms are sensitive to the statistical similarity of the images used in training and test datasets. The sensitivity of the base-layer segmentation algorithms affect the performances of the USF algorithm. Moreover, the employment of semi-supervision on the USF using Semi-supervised Segmentation Fusion algorithm further increase the performances.

Note that the performances of the proposed algorithms can be improved by the theoretical analyses on their open problems such as the investigation and modeling the dependency of the performances on the algorithm parameters, the statistical properties of the segmentations and images in training and test datasets, which are postponed to the future work. 

\section*{Acknowledgement}
This work was supported by the European commission project PaCMan EU FP7-ICT, 600918.

\bibliographystyle{IEEEtranS}
\bibliography{icip_r}

\begin{thebibliography}{10}
\providecommand{\url}[1]{#1}
\csname url@samestyle\endcsname
\providecommand{\newblock}{\relax}
\providecommand{\bibinfo}[2]{#2}
\providecommand{\BIBentrySTDinterwordspacing}{\spaceskip=0pt\relax}
\providecommand{\BIBentryALTinterwordstretchfactor}{4}
\providecommand{\BIBentryALTinterwordspacing}{\spaceskip=\fontdimen2\font plus
\BIBentryALTinterwordstretchfactor\fontdimen3\font minus
  \fontdimen4\font\relax}
\providecommand{\BIBforeignlanguage}[2]{{%
\expandafter\ifx\csname l@#1\endcsname\relax
\typeout{** WARNING: IEEEtranS.bst: No hyphenation pattern has been}%
\typeout{** loaded for the language `#1'. Using the pattern for}%
\typeout{** the default language instead.}%
\else
\language=\csname l@#1\endcsname
\fi
#2}}
\providecommand{\BIBdecl}{\relax}
\BIBdecl

\bibitem{gc1}
\BIBentryALTinterwordspacing
S.~Bagon, ``Matlab wrapper for graph cut,'' Dec 2006. [Online]. Available:
  \url{http://www.wisdom.weizmann.ac.il/~bagon}
\BIBentrySTDinterwordspacing

\bibitem{multispec}
L.~Biehl and D.~Landgrebe, ``Multispec: a tool for multispectral--hyperspectral
  image data analysis,'' \emph{Comput. and Geosci.}, vol.~28, pp. 1153--1159,
  Dec 2002.

\bibitem{admm}
S.~Boyd, N.~Parikh, E.~Chu, B.~Peleato, and J.~Eckstein, ``Distributed
  optimization and statistical learning via the alternating direction method of
  multipliers,'' \emph{Found. Trends Mach. Learn.}, vol.~3, no.~1, pp. 1--122,
  Jan 2011.

\bibitem{cut_seg}
Y.~Boykov and G.~Funka-Lea, ``Graph cuts and efficient n-d image
  segmentation,'' \emph{Int. J. Comput. Vision}, vol.~70, no.~2, pp. 109--131,
  Nov 2006.

\bibitem{gc3}
Y.~Boykov, O.~Veksler, and R.~Zabih, ``Efficient approximate energy
  minimization via graph cuts,'' \emph{IEEE Trans. Pattern Anal. Mach.
  Intell.}, vol.~20, no.~12, pp. 1222--1239, Nov 2001.

\bibitem{semibook}
O.~Chapelle, B.~Sch{\"o}lkopf, and A.~Zien, Eds., \emph{Semi-Supervised
  Learning}.\hskip 1em plus 0.5em minus 0.4em\relax Cambridge, MA: MIT Press,
  2006.

\bibitem{ms}
D.~Comaniciu and P.~Meer, ``Mean shift: A robust approach toward feature space
  analysis,'' \emph{IEEE Trans. Pattern Anal. Mach. Intell.}, vol.~24, no.~5,
  pp. 603--619, May 2002.

\bibitem{o3}
B.~Dasarathy, \emph{Decision fusion}.\hskip 1em plus 0.5em minus 0.4em\relax
  IEEE Computer Society Press, 1994.

\bibitem{for}
D.~A. Forsyth and J.~Ponce, \emph{Computer Vision: A Modern Approach}.\hskip
  1em plus 0.5em minus 0.4em\relax Prentice Hall Professional Technical
  Reference, 2002.

\bibitem{firfir}
L.~Franek, D.~D. Abdala, S.~Vega-Pons, and X.~Jiang, ``Image segmentation
  fusion using general ensemble clustering methods,'' in \emph{Proceedings of
  ACCV'10}, ser. ACCV'10, 2011, pp. 373--384.

\bibitem{BOEM}
A.~Goder and V.~Filkov, ``Consensus clustering algorithms: Comparison and
  refinement,'' in \emph{Proc. SIAM Workshop on Algorithm Engineering and
  Experiments}, J.~I. Munro and D.~Wagner, Eds., 2008, pp. 109--117.

\bibitem{gonz}
R.~C. Gonzalez and R.~E. Woods, \emph{Digital Image Processing}, 2nd~ed.\hskip
  1em plus 0.5em minus 0.4em\relax Boston, MA, USA: Addison-Wesley Longman
  Publishing Co., Inc., 2001.

\bibitem{w1}
T.~Li and C.~H.~Q. Ding, ``Weighted consensus clustering,'' in \emph{SIAM Int.
  Conf. on Data Mining}, Atlanta, Georgia, 2008, pp. 798--809.

\bibitem{sf}
M.~Ozay, F.~Yarman~Vural, S.~Kulkarni, and H.~Poor, ``Fusion of image
  segmentation algorithms using consensus clustering,'' in \emph{Proc. of Int.
  Conf. Image Processing, ({ICIP} 2013)}, Sep 2013, pp. 4049 -- 4053.

\bibitem{porway}
J.~Porway, Q.~Wang, and S.~C. Zhu, ``A hierarchical and contextual model for
  aerial image parsing,'' \emph{Int. J. Comput. Vision}, vol.~88, no.~2, pp.
  254--283, Jun 2010.

\bibitem{fl3}
V.~Sharma and J.~Davis, ``Feature-level fusion for object segmentation using
  mutual information,'' in \emph{Augmented Vision Perception in Infrared}, ser.
  Advances in Pattern Recognition, R.~Hammoud, Ed.\hskip 1em plus 0.5em minus
  0.4em\relax Springer London, 2009, pp. 295--320.

\bibitem{nc1}
J.~Shi and J.~Malik, ``Normalized cuts and image segmentation,'' \emph{IEEE
  Trans. Pattern Anal. Mach. Intell.}, vol.~22, pp. 888--905, Aug 2000.

\bibitem{lasso}
R.~Tibshirani, ``Regression shrinkage and selection via the lasso,'' \emph{J.
  Roy. Stat. Soc. B}, vol.~58, pp. 267--288, 1996.

\bibitem{Info_theoretic}
N.~X. Vinh, J.~Epps, and J.~Bailey, ``Information theoretic measures for
  clusterings comparison: is a correction for chance necessary?'' in
  \emph{Proc. of Int. Conf. Machine Learning (ICML)}, 2009, pp. 1073--1080.

\bibitem{convex2}
F.~Wang, X.~Wang, and T.~Li, ``Generalized cluster aggregation,'' in \emph{Proc
  of IJCAI}, 2009, pp. 1279--1284.

\bibitem{lhi}
B.~Yao, X.~Yang, and S.-C. Zhu, ``Introduction to a large-scale general purpose
  ground truth database: methodology, annotation tool and benchmarks,'' in
  \emph{Proc. Int. Conf. Energy Minimization Comput. Vis. Pattern Recognit.},
  2007, pp. 169--183.

\end{thebibliography}
\end{document}